%% file: iclr2025_conference.tex
\documentclass[table,xcdraw]{article} 
\usepackage{iclr2025_conference,times}

\input{math_commands.tex}

\usepackage{amsmath,amsfonts}
\usepackage{hyperref}
\usepackage{url}
\usepackage{enumitem}  
\usepackage{multirow}
\usepackage{graphicx}        
\usepackage{booktabs}       
\usepackage{nicefrac}       
\usepackage{microtype}      
\usepackage{pifont}
\usepackage{arydshln}
\usepackage[misc]{ifsym}
\usepackage{wrapfig}

\title{\textbf{Co$^{\mathbf{3}}$Gesture:} Towards \underline{Co}herent \underline{Co}ncurrent \underline{Co}-speech 3D Gesture Generation with Interactive Diffusion}

\author{Xingqun Qi$^{1, }$\footnotemark[1], Yatian Wang$^{1, }$\footnotemark[1], Hengyuan Zhang$^{2}$, Jiahao Pan$^{1}$\\
\textbf{Wei Xue$^{1}$, Shanghang Zhang$^{2}$, Wenhan Luo$^{1}$, Qifeng Liu$^{1, \textrm{\Letter}}$, Yike Guo$^{1, \textrm{\Letter}}$}\\
$^{1}$ The Hong Kong University of Science and Technology\\  $^{2}$ Peking University 
}

\newcommand{\ie}{\emph{i.e.}}
\newcommand{\eg}{\emph{e.g.}}
\newcommand{\wrt}{\emph{w.r.t.}}

\newcommand{\etc}{\emph{etc. }}
\newcommand{\modelname}{Co$^{3}$Gesture }
\iclrfinalcopy 
\begin{document}
{
\maketitle
\begin{figure}[h]
\centering
\vspace{-2.5em}
\includegraphics[width=1\textwidth]{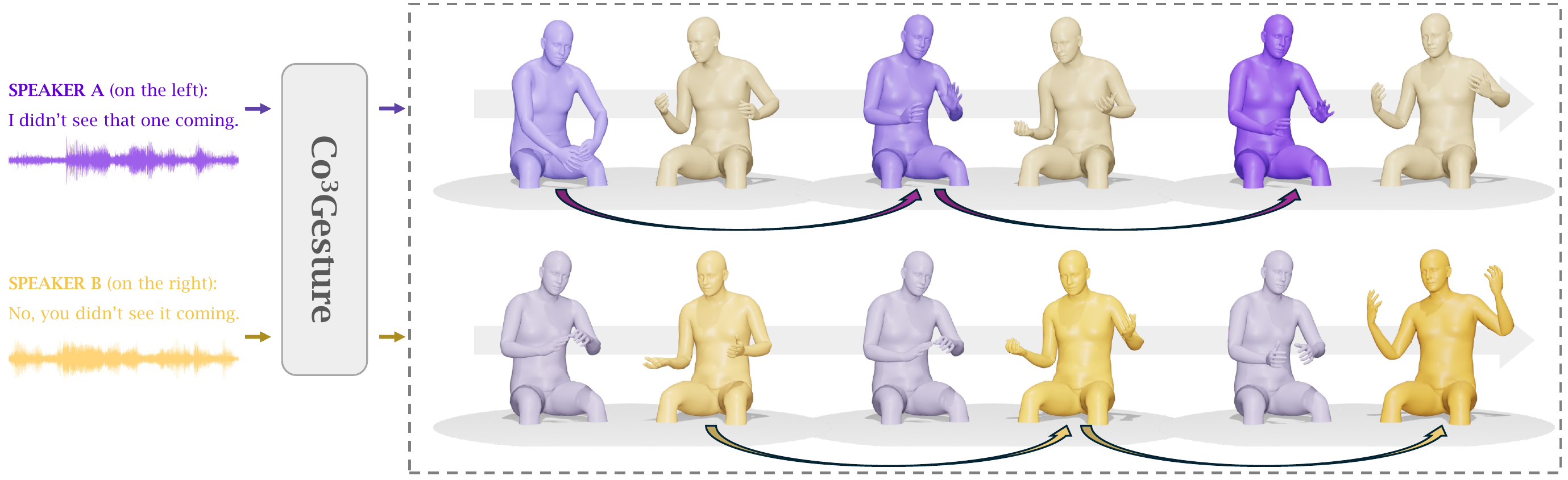}
\vspace{-1.5em}
\caption{Diverse exemplary clips sampled by our method from our newly collected \textbf{GES-Inter dataset}. The vital frames are visualized to demonstrate the concurrent upper body dynamics of two speakers generated by our \textbf{\modelname} framework displaying temporal coherent interaction with each other, respectively. Best view on screen.}

%
\label{fig:teaser}
\end{figure}}

\footnotetext[1]{These authors contributed equally to this work.}
\footnotetext{$\textrm{\Letter}$ Corresponding authors.}

\input{section/0_abstract}    
\input{section/1_intro}
\input{section/2_related_work}
\input{section/3_methodology}

\input{section/4_experiments}
\input{section/5_conclusion}


\bibliography{iclr2025_conference}
\bibliographystyle{iclr2025_conference}


\newpage
\input{section/6_appendix}

\end{document}

%% file: math_commands.tex
\usepackage{amsmath,amsfonts, bm}

\def\eqref#1{equation~\ref{#1}}

\def\1{\bm{1}}

\DeclareMathAlphabet{\mathsfit}{\encodingdefault}{\sfdefault}{m}{sl}
\SetMathAlphabet{\mathsfit}{bold}{\encodingdefault}{\sfdefault}{bx}{n}



%% file: section/0_abstract.tex
\begin{abstract}
Generating gestures from human speech has gained tremendous progress in animating virtual avatars.
While the existing methods enable synthesizing gestures cooperated by individual self-talking, they overlook the practicality of concurrent gesture modeling with two-person interactive conversations.
Moreover, the lack of high-quality datasets with concurrent co-speech gestures also limits handling this issue. 
To fulfill this goal, we first construct a large-scale concurrent co-speech gesture dataset that contains more than 7M frames for diverse two-person interactive posture sequences, dubbed \textbf{GES-Inter}.
Additionally, we propose \textbf{Co$^3$Gesture}, a novel framework that enables coherent concurrent co-speech gesture synthesis including two-person interactive movements. 
Considering the asymmetric body dynamics of two speakers, our framework is built upon two cooperative generation branches conditioned on separated speaker audio.
Specifically, to enhance the coordination of human postures \wrt corresponding speaker audios while interacting with the conversational partner, we present a Temporal Interaction Module (\textbf{TIM}). 
TIM can effectively model the temporal association representation between two speakers' gesture sequences as interaction guidance and fuse it into the concurrent gesture generation. 
Then, we devise a mutual attention mechanism to further holistically boost learning dependencies of interacted concurrent motions, thereby enabling us to generate vivid and coherent gestures.
Extensive experiments demonstrate that our method outperforms the state-of-the-art models on our newly collected GES-Inter dataset.
The dataset and source code are publicly available at \href{https://mattie-e.github.io/Co3/}{\textit{https://mattie-e.github.io/Co3/}}.

\end{abstract}

%% file: section/1_intro.tex
\section{Introduction}
The generation of co-speech gestures seeks to create expressive and diverse human postures that align with audio input. These non-verbal behaviors play a crucial role in human communication, significantly enhancing the effectiveness of speech delivery.
Meanwhile, modeling co-speech gestures has broad applications in embodied AI, including human-machine interaction~\citep{liu2023audio}, robotic assistance~\citep{farouk2022studying}, and virtual/augmented reality (AR/VR)~\citep{fu2022systematic}. Traditionally, researchers have primarily concentrated on synthesizing upper body gestures that correspond to spoken audio~\citep{liu2022learning, yi2023generating}.

These methods usually focus on synthesizing single-speaker gestures following people's self-talking~\citep{liu2024towards, Liu_2024_CVPR, Qi_2024_CVPR, yang2023diffusestylegesture}. Although some researchers model the single human postures via conversational speech corpus~\citep{Mughal_2024_CVPR,
Ng_2024_CVPR}, they mostly overlook generating the concurrent long sequence gestures with interactions. Besides, others generate the single speaker gesture from conversational corpus incorporated with interlocuter reaction movements~\citep{kucherenko2023genea, zhao2023diffugesture}.
However, few researchers have devoted themselves to constructing datasets with interactive concurrent gestures.
For example, the interactive movement of two speakers may include waving arms when saying ``\emph{hello}'' during conversation. In this work, we therefore introduce the new task of two-speaker concurrent gestures generation under the condition of conversational human speeches, as displayed in Figure~\ref{fig:teaser}. 

There are two main challenges in this task: 1) Datasets of concurrent 3D co-speech gestures synchronized with conversation audios of two speakers are scarce. Creating such a dataset containing large-scale 3D human postures is difficult due to complex motion capture systems and expensive labor for actors. 2) Modeling the plausible and temporal coherent co-speech gestures of two speakers is difficult, especially involving the frequent interactions in long sequences.

To address the issue of data scarcity, we construct a new large-scale whole-body meshed 3D co-speech gesture dataset that includes concurrent speaker postures within more than 7M frames, dubbed \textbf{GES-Inter}. In particular, we first leverage the advanced 3D pose estimator~\citep{zhang2023pymaf}
\begin{wrapfigure}{r}{0.5\textwidth}
    \centering
\includegraphics[width=0.5\textwidth]{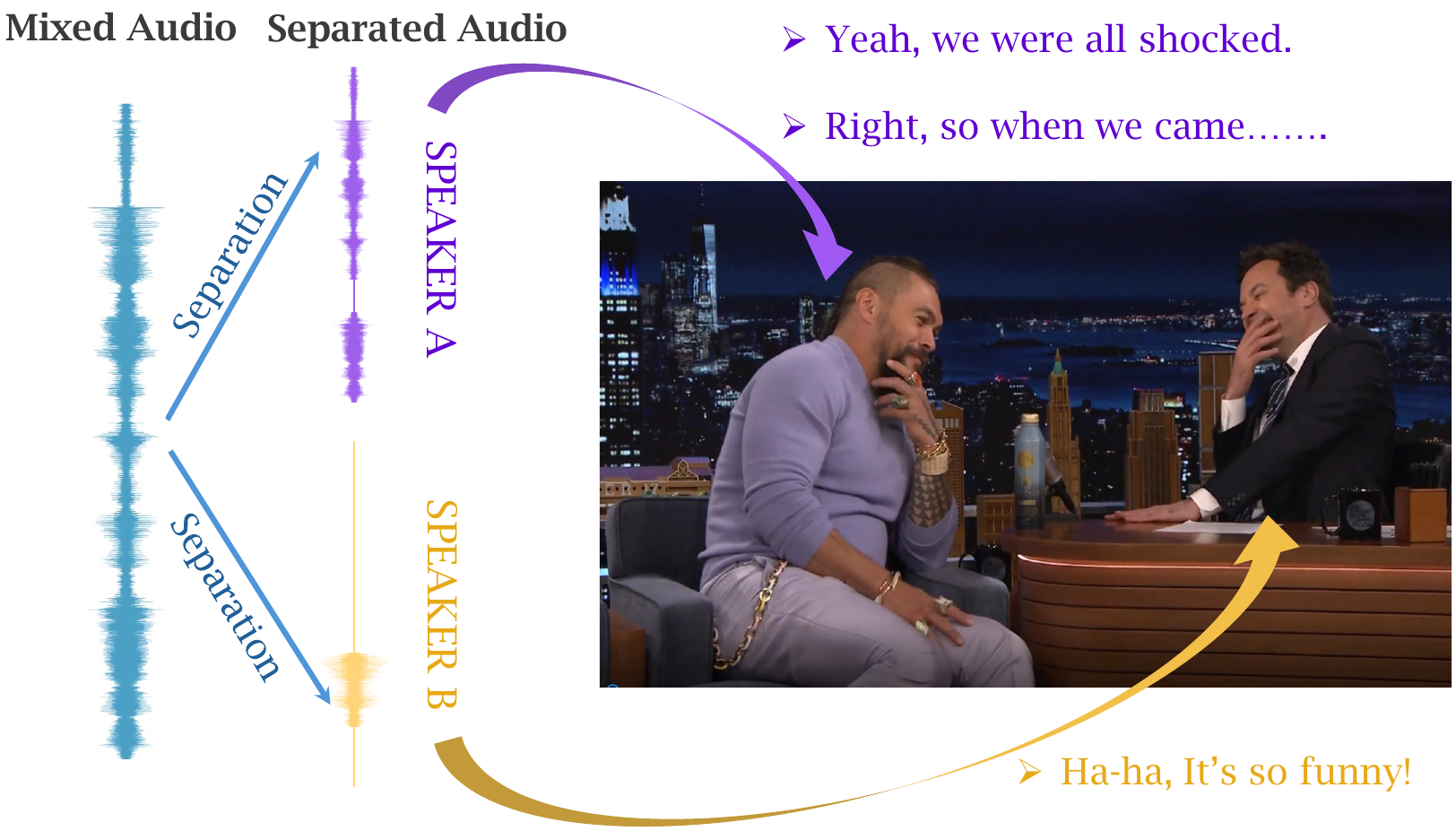}
\caption{Illustration of our audio separation and alignment with speakers.}
    \label{fig:dataset}
\end{wrapfigure}
to obtain high-quality poses (\ie, SMPL-X~\cite{pavlakos2019expressive} and FLAME~\cite{li2017learning}) from in-the-wild talk show videos. To obtain the individual sound signals of each speaker in the conversation while preserving the identity consistency with the posture movement, we employ the pyannote-audio~\cite{bredin2020pyannote} to separate the mixed speech, as shown in Figure~\ref{fig:dataset}. Afterward, by utilizing the automatic speech recognition techniques Whisper-X~\cite{bain2022whisperx}, we acquire the consistent text transcript and speech phoneme with speaker audio. In this fashion, our GES-Inter dataset covers a wide range of two-person interactive concurrent co-speech gestures, from daily conversations to formal interviews. Moreover, the multi-modality annotation and common meshed human postures pave the potential for various downstream tasks like human behaviors analysis~\citep{liang2024intergen, xu2024inter} and talking face generation~\citep{peng2024synctalk, Ng_2024_CVPR}, \etc 

Based on our GES-Inter dataset, we propose a novel framework, named \textbf{\modelname}, to model the coherent concurrent co-speech 3D gesture generation. The key insight of our framework is to carefully build the interactions between concurrent gestures. Here, we observe that the motions of two speakers are asymmetric (\eg, when one speaker moves in talking, the other could be quiet in static or moving slowly). Directly producing the concurrent gestures in a holistic manner may lead to unnatural results. Therefore, we establish two cooperative transformer-based diffusion branches to generate corresponding gestures of two speakers, performing the specific denoising process, respectively. This bilateral paradigm encourages our framework to yield diverse interactive movements while effectively preventing mode collapse.

Moreover, to ensure the motions of the one speaker are temporally consistent with the corresponding audio signal and display coherent interaction with the conversational partner, we devise a Temporal Interaction Module (\textbf{TIM}). Specifically, we first incorporate the separated human voices to produce single-speaker gesture features, respectively. Then, we model the joint embedding of the current speaker features and the integrated conversational motion clues guided by mixed speech audio. Here, the learned joint embedding is leveraged as the soft weight to balance the interaction dependence of the generated current speaker gesture dynamics with other ones. Then, we conduct the mutual attention of the fused bilateral gesture denoisers to further facilitate high-fidelity concurrent gesture generation with desirable interactive properties. Extensive experiments conducted on our newly collected GES-Inter dataset verify the effectiveness of our method, displaying diverse and vivid concurrent co-speech gestures.

\noindent Overall, our contributions are summarized as follows:
\begin{itemize}[leftmargin=*]
\item We introduce the new task of concurrent co-speech gesture generation cooperating with one newly collected large-scale dataset named GES-Inter. It contains more than 7M high-quality co-speech postures of the whole body, significantly facilitating research on diverse gesture generation.

\item We propose a novel framework named \modelname upon the bilateral cooperative diffusion branches to produce realistic concurrent co-speech gestures. Our \modelname includes the tailor-designed Temporal Interaction Module (TIM) to ensure the temporal synchronization of gestures \wrt the corresponding speaker voices while preserving desirable interactive dynamics.

\item Extensive experiments show that our framework outperforms various state-of-the-art counterparts on the GES-Inter dataset, producing diverse and coherent concurrent co-speech gestures given conversational speech audios.

\end{itemize}

%% file: section/2_related_work.tex
\section{Related Work}

\paragraph{Co-speech Gesture Synthesis.}
Synthesizing the diverse and impressive co-speech gestures displays a significant role in the wide range of applications like human-machine interaction~\citep{cho2023real, guo2021human}, robot~\citep{de2023design, sahoo2023hand}, and embodied AI~\citep{li2023understanding,benson2023embodied}. Numerous works were proposed to address this task that can be roughly divided into rule-based approaches, machine learning designed methods, and deep learning based ones. Rule-based research depends on linguistic experts' pre-defined corpus to bridge human speech and gesture movements~\citep{cassell1994animated, poggi2005greta}. The others usually leverage machine learning techniques with mutually constructed speech features to generate co-speech gestures~\citep{levine2010gesture, sargin2008analysis}. However, these methods heavily rely on efforts on pre-processing which may cause expensive labor consumption.

Recently deep learning based methods gained much development directly modeling co-speech gesture synthesis via deep neural networks. Most of them usually leverage the multi-modality cues to generate postures incorporated with individual self-talking audio~\citep{Li_2021_ICCV, zhu2023taming, yi2023generating, qi2024cocogesture}, such as speaker identity~\citep{Liu_2024_CVPR, liu2024towards, chen2024diffsheg}, emotion~\citep{Qi_2024_CVPR, qi2023emotiongesture, liu2022beat}, and transcripts~\citep{zhang2024semantic, ao2023gesturediffuclip, ao2022rhythmic, zhi2023livelyspeaker}. Only a few counterparts propose to synthesize the single gesture under conversational speech guidance~\citep{Ng_2024_CVPR, mughal2024convofusion}. Besides, the GENEA challenge holds the most similar settings to us. The participants aim to generate the single-person gesture from the conversational corpus incorporated with interlocuter reaction movements~\citep{kucherenko2023genea}.
However, they overlook the concurrent co-speech gesture modeling of two speakers during the conversation is much more practical in the real scenes. Few of the above methods could be directly adapted to this new thought.

\paragraph{Co-speech Gesture Datasets.}
Co-speech gesture datasets are roughly divided into two types: pseudo-label based gestures and motion-capture based ones. For pseudo-label approaches, researchers usually utilize the pre-trained pose estimator to obtain upper body postures from in-the-wild News or talk show videos~\citep{yoon2020speech, ahuja2020style, habibie2021learning}. Thanks to the recent advanced parametric whole-body meshed 3D model SMPL-X~\cite{pavlakos2019expressive} and FLAME~\cite{li2017learning}, some high-quality whole-body based 3D co-speech gesture datasets are emerged~\citep{qi2024cocogesture,yi2023generating,Qi_2024_CVPR,qi2023emotiongesture}. Meanwhile, it significantly promotes the construction of motion-capture based co-speech datasets~\citep{liu2022beat, Liu_2024_CVPR, ghorbani2023zeroeggs, mughal2024convofusion, Ng_2024_CVPR}. Although some of them are built upon conversational corpora, they only provide gestures of single speakers~\citep{Liu_2024_CVPR, Ng_2024_CVPR}. The TWH16.2~\citep{lee2019talking} dataset displays the pioneer exploration of concurrent gestures via joint-based representation. However, it overlooks the significance of the facial expression data in conversation. Meanwhile, the SMPL-X mesh-based whole-body data in our dataset is more convenient for avatar rendering and downstream applications (\eg, talking face) compared to TWH16.2.
Besides, the DND GROUP GESTURE dataset~\cite{mughal2024convofusion} is built upon a multi-performer group talking scene, which can not be directly applied to our task.
Therefore, a 3D co-speech dataset including concurrent gestures of two speakers with the meshed whole body is required for further research.

\paragraph{3D Human Motion Modeling.}
Human motion modeling aims to generate natural and realistic coherent posture sequences from multi-modality conditions, which contains co-speech gesture synthesis as a sub-task~\citep{liang2024omg, tevet2023human}. One of the hottest tasks is generating human movements from the input action descriptions~\citep{jiang2023motiongpt, zhang2023generating, lin2024motion, xu2024inter}. It needs to enforce the results by displaying an accurate semantic expression aligned with text prompts. The other one that shares modality guidance similar to our task is the AI choreographer~\citep{li2020learning, li2021ai, siyao2022bailando, le2023controllable}. 
While retaining analogous interactive human motion modeling with the approaches mentioned above, our newly introduced work differs from them significantly. Both of the aforementioned topics follow the symmetrical fact that exchanging the identities of performers during interactions does
not change the semantics or coherence of motions.
We take the asymmetric body dynamics of concurrent human movements into consideration, motivating us to design the bilateral diffusion branches.

%% file: section/3_methodology.tex
\section{Proposed Method}

\subsection{Interactive Gesture Dataset Construction}
\label{sec3.1}

\paragraph{Preliminary.} 
Due to the expensive labor and complex motion capture system establishment during the frequent interactive conversations, similar to \cite{yi2023generating, liu2022learning, qi2024cocogesture}, we intend to obtain the high-quality 3D pseudo human postures of our dataset.  Synthesizing datasets conducive to our task focuses on ensuring high-fidelity and smooth gesture movements, authority speaker voice separation, and identity consistent audio-posture alignment.

\begin{table}[t]
\centering
\caption{Statistical comparison of our GES-Inter with existing datasets.  The dotted line separates whether the speech content in the dataset is built based on the conversational corpus. }
\label{tab:dataset}
\resizebox{\textwidth}{!}{%
\begin{tabular}{lccccccc}
\toprule
\multirow{2}{*}{Datasets} & \multirow{2}{*}{\begin{tabular}[c]{@{}c@{}}Concurrent\\ Gestures\end{tabular}} & \multirow{2}{*}{\begin{tabular}[c]{@{}c@{}}Duration\\ (hours)\end{tabular}} & \multicolumn{4}{c}{Additional Attributes} & \multirow{2}{*}{\begin{tabular}[c]{@{}c@{}}Joint\\ Annotation\end{tabular}} \\ \cmidrule(r){4-7}
 &  &  & Facial & Mesh & Phonme & Text &  \\ \midrule \midrule
TED~\citep{yoon2020speech}\textcolor[HTML]{C0C0C0}{$_{TOG}$} & \ding{55} & 106.1 & \ding{55} & \ding{55} & \ding{55} & \ding{51} & pseudo \\ 
TED-Ex~\citep{liu2022learning}\textcolor[HTML]{C0C0C0}{$_{CVPR}$} & \ding{55} & 100.8 & \ding{55} & \ding{55} & \ding{55} & \ding{51} & pseudo \\
EGGS~\citep{ghorbani2023zeroeggs}\textcolor[HTML]{C0C0C0}{$_{CGF}$} & \ding{55} & 2 & \ding{55} & \ding{55} & \ding{55} & \ding{51} & pseudo \\
BEAT~\citep{liu2022beat}\textcolor[HTML]{C0C0C0}{$_{ECCV}$} & \ding{55} & 76 & \ding{51} & \ding{55} & \ding{51} & \ding{51} & mo-cap \\
SHOW~\citep{yi2023generating}\textcolor[HTML]{C0C0C0}{$_{CVPR}$} & \ding{55} & 26.9 & \ding{51} & \ding{51} & \ding{55} & \ding{55} & pseudo \\ 
\hline
{TWH16.2~\citep{lee2019talking}\textcolor[HTML]{C0C0C0}{$_{ICCV}$} }& \ding{51} & 17 & \ding{51} & \ding{51} & \ding{51} & \ding{51} & mo-cap \\
BEAT2~\citep{Liu_2024_CVPR}\textcolor[HTML]{C0C0C0}{$_{CVPR}$} & \ding{55} & 60 & \ding{51} & \ding{51} & \ding{51} & \ding{51} & mo-cap \\
DND~\citep{mughal2024convofusion}\textcolor[HTML]{C0C0C0}{$_{CVPR}$} & \ding{55} & 6 & \ding{55} & \ding{55} & \ding{55} & \ding{55} & mo-cap \\
Photoreal~\citep{Ng_2024_CVPR}\textcolor[HTML]{C0C0C0}{$_{CVPR}$} & \ding{55} & 8 & \ding{51} & \ding{55} & \ding{55} & \ding{55} & mo-cap \\ \midrule
\rowcolor[HTML]{ECF4FF}
\textbf{GES-Inter (ours)} & \ding{51} & \textbf{70} & \ding{51} & \ding{51} & \ding{51} & \ding{51} & \textbf{pseudo} \\ \bottomrule
\end{tabular}%
}
\end{table}

\paragraph{Estimation of 3D Posture.} 
Firstly, we exploit the state-of-the-art 3D pose estimator Pymaf-X~\cite{zhang2023pymaf} to acquire the meshed whole-body parametric human postures based on SMPL-X~\cite{pavlakos2019expressive}. In particular, the body dynamics are denoted by the unified SMPL model~\cite{loper2023smpl} that collaborated with the MANO hand model~\cite{boukhayma20193d}. Meanwhile, we adopt the FLAME face model~\cite{li2017learning} to present the facial expressions of speakers.
The corpora are collected from the in-the-wild talk show or formal interview videos that contain high-resolution frames and unobstructed sitting postures. Then, we conduct extensive data processing to filter the unnatural and jittery poses, thereby ensuring the high-quality of the dataset\footnote{Please refer to supplementary material for more details about data processing.}. Our GES-Inter includes more than 7M validated gesture frames with 70 hours. To the best of our knowledge, this is the first large-scale co-speech dataset that includes mesh-based whole-body concurrent postures of two speakers, as reported in Table~\ref{tab:dataset}.

\paragraph{Separation of Speaker Audio.} 
To obtain the identity-specific speech audios of each speaker in the mixed conversation corpus, we leverage the advanced sound source separation technique pyannote-audio~\cite{bredin2020pyannote} to conduct human voice separation. Here, we enforce the number of separated speakers as two for assigning each speech segment to the corresponding speaker. 
Then, we utilize the superior speech recognition model WhisperX~\cite{bain2022whisperx} to acquire accurate word-level text transcripts. Once we acquire high-fidelity transcripts, we utilize the Montreal Forced Aligner (MFA)~\cite{mcauliffe2017montreal} to obtain phoneme-level timestamps associated with facial expressions. Such extensive multi-modality attributes of our dataset enable the research of various downstream tasks like talking face~\citep{peng2024synctalk, Ng_2024_CVPR}, and human behavior analysis~\citep{park2023generative, qi2023diverse, dubois2024alpacafarm} \etc

\paragraph{Alignment of Audio-Posture Pair.} 
Once we obtain the separated audio signals for each speaker, we assign them to the corresponding body dynamics. We recruit professional inspectors to manually annotate the separated audio signals with their corresponding speaker identities. To ensure the accuracy of the aligned audio-posture pairs, different inspectors double-check these results. In this way, our newly constructed GES-Inter dataset offers high-quality concurrent gestures with aligned multi-modal annotations.

\subsection{Problem Formulation}
\label{sec3.2}

Given a sequential collection of conversational audio signal $\mathbf{C}_{mix}$ as the condition, our goal is to generate the interactive concurrent gesture sequences of two speakers $\mathbf{x}$. Where $\mathbf{C}_{mix} = \mathrm{C}_{a} + \mathrm{C}_{b}$
and $\mathbf{x} = \left \{ \mathrm{x}_{a}, \mathrm{x}_{b} \right \}$ denote the corresponding audios and postures of two speakers. The sequence length is the fixed number $N$. Specifically, each pose of a single person is presented as $J$ joints with 3D representation. Note that we only generate the upper body including fingers in this work.

\begin{figure}[t]
\begin{center}
\includegraphics[width=1\linewidth]{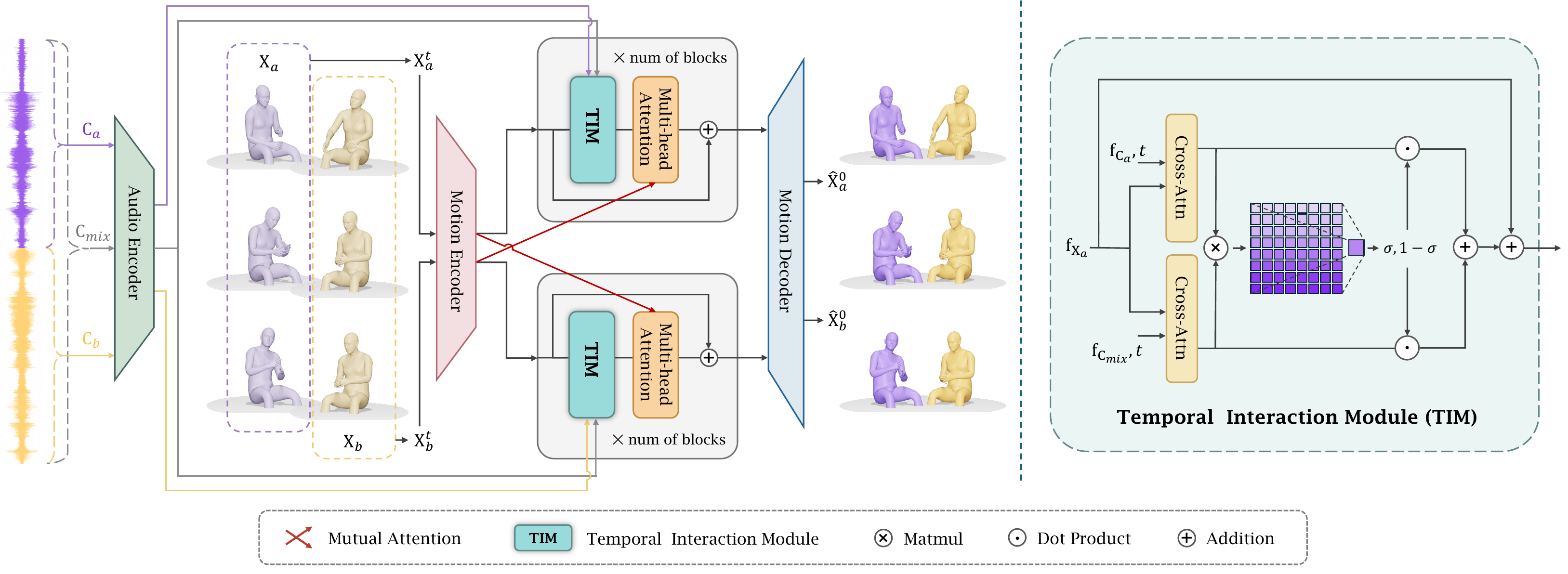}
\end{center}
\caption{The overall pipeline of our \modelname. Given conversational speech audios, our framework generates concurrent co-speech gestures with coherent interactions.
}
\label{fig:pipeline}
\end{figure}

\subsection{Bilateral Cooperative Diffusion Model}
\label{sec3.3}

Considering the asymmetric body dynamics of two speakers, we aim to address the concurrent co-speech gesture generation in a bilateral cooperation manner, as depicted in Figure~\ref{fig:pipeline}. The framework takes the two noisy human motions as input for producing denoised ones, which is conditioned on diffusion time step $t$, mixed conversational audio signal $\mathbf{C}_{mix}$, and separated speaker voices $\mathrm{C}_{a}$ or $\mathrm{C}_{b}$. 
We leverage separated human speech as guidance for bilateral branches to generate corresponding gestures. Moreover, we utilize the original mixed audio signal of two speakers to indicate the interaction information to ensure the synthesized posture retains rhythm with specific audio while preserving interactive coherency with the conversation partner.
All the audio signals are fed into the audio encoder for feature extraction.

\paragraph{Temporal Interaction Module.}
To ensure temporal consistency while preserving the interactive dynamics of concurrent gestures, we propose a Temporal Interaction Module (TIM) to model the temporal association representation between each current speaker movements and conversational counterparts. As shown in Figure~\ref{fig:pipeline}, we utilize the features extracted from mixed conversational audios to indicate the interaction information. Here, the dimensions of features in our TIM are all normalized as $\mathbb{R}^{N\times D}$. For notation simplicity, we take single-branch $\mathrm{x}_{a}$ for elaboration.

In particular, we first incorporate the current speaker audio embedding $\mathbf{f}_{\mathrm{C}_{a}}$ as the query $Q$ to match the key feature $K$ and value feature $V$ belonging to motion embedding $\mathbf{f}_{\mathrm{x}_{a}}$ via cross-attention meshanism~\citep{vaswani2017attention}:
\begin{align}
Q = \mathbf{f}_{\mathrm{C}_{a}} \mathbf{W},  K = {\mathbf{f} _{\mathrm{x}_{a}}} \mathbf{W}, V = {\mathbf{f} _{\mathrm{x}_{a}} } \mathbf{W}.
\label{eq3}
\end{align} 
Here, $\mathbf{W}$ denotes the projection matrix. 
Along with this operation, we obtain the updated current speaker motion embedding $\mathbf{f} _{\mathrm{x}_{a}, \mathrm{C}_{a}}$. Similarly, we acquire the interactive motion embedding $\mathbf{f} _{\mathrm{x}_{a}, \mathbf{C}_{mix}}$ incorporated with mixed conversational speeches. Then we calculate the temporal correlation matrix $\mathbf{M}\in \mathbb{R}  ^{N\times N}$ between the updated current gesture embedding and interactive embedding. Here, the temporal correlation matrix represents the temporal variants between the current gesture sequences and interactive ones. Then, we exploit a motion encoder to acquire a learnable weight parameter $\sigma$ as the temporal-interaction dependency. Once we obtain the weight parameter, the current speaker motion embedding is boosted as follows:
\begin{align}
\mathbf{f} _{\mathrm{x}_{a}, \mathrm{C}_{a}} = \sigma  \odot \mathbf{f} _{\mathrm{x}_{a},  \mathrm{C}_{a}} + (1-\sigma)\odot \mathbf{f} _{\mathrm{x}_{a}, \mathbf{C}_{mix}}, \sigma  = sigmoid(\text{Enc}(\mathbf{M})) ,
\end{align}
where $\odot$ is Hadamard product, \text{Enc} denotes the motion encoder. The motion embedding of the conversation partner is updated in the same manner. In this fashion, the temporal interaction fidelity of generated gestures is well-preserved.

\vspace{-0.75em}
\paragraph{Mutual Attention Mechanism.}
To further enhance the interaction between two speakers, we construct bilateral cooperative branches that interact with each other to produce concurrent gestures. To be specific, we introduce the mutual attention layers that take the features of the counterpart as the query $Q$ in Multi-Head Attention (MHA), respectively. 
We observe that exchanging the input order of the speaker's audio results in an invariance effect of interactive body dynamics. In other words, the distribution of interaction data of two speakers adheres to the same marginal distribution. 
Therefore, we formulate the cooperating denoisers retaining shared weight update strategies.
This encourages the gesture features after the TIM to be more temporal and interactive with partner ones, holistically.

\subsection{Objective Functions}
\label{sec3.4}

During the training phase, the denoisers of our bilateral branches share the common network structure. Given the diffusion time step $t$, the current speaker audio $\left \{ \mathrm{C}_{a}, \mathrm{C}_{b} \right \}$, the mixed conversation audio $\mathbf{C}_{mix}$, and the noised gestures $\left \{\mathrm{x}_{a}^{(t)}, \mathrm{x}_{b}^{(t)} \right \}$, the denoisers are enforced to produce continuous human gestures. The denoising process can be constrained by the simple objective:
\begin{align}
\mathcal{L}_{simple} & = \mathbb{E} _{\mathbf{x},t,\epsilon }  \left [\left \| \mathrm{x}_{a}-\mathcal{D}(\mathrm{x}_{a}^{(t)},\mathrm{C}_{a}, \mathbf{C}_{mix},t)\right \|_{2}^{2}  +  \left \| \mathrm{x}_{b}-\mathcal{D}(\mathrm{x}_{b}^{(t)},\mathrm{C}_{b}, \mathbf{C}_{mix},t)\right \|_{2}^{2} \right ],
\label{eq1}
\end{align}
where $\mathcal{D}$ is the denoiser, $\epsilon\sim \mathcal{N}(\mathbf{0} , \mathbf{I})$ is the added random Gaussian noise, $\mathrm{x}_{\left \{ a, b \right \}}^{(t)} = \mathrm{x}_{\left \{ a, b \right \}} + \sigma _{(t)}\epsilon$ is the gradually noise adding process at step $t$. $\sigma _{(t)}\in   (0, 1)$ is the constant hper-parameter. Moreover, we utilize the velocity loss $\mathcal{L}_{vel}$ and foot contact loss $\mathcal{L}_{foot}$ ~\citep{tevet2023human} to provide supervision on the smoothness and physical reasonableness, respectively. Finally, the overall objective is:
\begin{align}
\mathcal{L}_{total} & = \lambda _{simple} \mathcal{L}_{simple} + \mathcal{L}_{vel} +  \mathcal{L}_{foot},
\label{eq2}
\end{align}
where $\lambda _{simple}$ is trade-off weight coefficients.

In the inference, since the audio signals of the concurrent gestures generation serve as an essential condition modality, the prediction of human postures is formulated as fully conditioned denoising. This encourages our framework to strike a balance between high fidelity and diversity.

%% file: section/4_experiments.tex
\section{Experiments}
\subsection{Datasets and Experimental Setting}
\paragraph{GES-Inter Dataset.}
Since the existing co-speech gesture datasets fail to provide interactive concurrent body dynamics, we contribute a new dataset named GES-Inter to evaluate our approach. The human postures of our GES-Inter are collected from 1,462 processed videos including talk shows and formal interviews. The extraction takes 8 NVIDIA RTX 4090 GPUs in one month, obtaining 20 million raw frames. After the complex data processing, we get more than 7 million validated instances. 
Finally, we acquire 27,390 motion clips that are split into training/ validation/ testing following criteria~\citep{liu2022beat,Liu_2024_CVPR} as 85\%/ 7.5\%/ 7.5\%.

\paragraph{Implementation Details.} 

We set the total generated sequence length $N=90$ with the FPS normalized as 15 in the experiments. 
$\mathbf{C}_{mix}$, $\mathrm{C}_{a}$, and $\mathrm{C}_{b}$ are represented as audio signal waves, initially. Then,
these audio signals are converted into mel-spectrograms with an FFT window size of 1,024, and the hop length is 512. The dimension of input audio mel-spectrograms is $128 \times 186$.
We follow the tradition of ~\citep{liu2022learning, Qi_2024_CVPR, qi2024cocogesture} to leverage the speech recognizer as the audio encoder. Each branch of our pipeline is implemented with 8 blocks within 8 heads of attention layers. The latent dimension $D$ is set to 768. 

In the training stage, we set $\lambda _{simple} = 15$, empirically. The initial learning rate is set as $1\times 10^{-4}$ with an AdamW optimizer. Similar to \cite{nichol2021improved}, we set the diffusion time step as 1,000 with the cosine noisy schedule. Our model is applied on a single NVIDIA H800 GPU with a batch size of 128. The training takes a total of 100 epochs, accounting for 3 days. During inference, we adopt DDIM~\cite{song2020denoising} sampling strategy with 50 denoising timesteps to produce gestures. Our experiments only contain upper body joints without facial expressions and shape parameters. Our \modelname synthesizes upper body movements containing 46 joints (\ie, 16 body joints + 30 hand joints) of each speaker. Each joint is converted to a 6D rotation representation~\cite{zhou2019continuity} for more stable modeling. The dimension of the generated motion sequence is $\mathbb{R}^{90\times 276}$, where 90 denotes frame number and $276 = 46 \times 6$ means upper body joints. The order of each joint follows the original convention of SMPL-X.

\paragraph{Evaluation Metrics.} 
To fully evaluate the realism and diversity of the generated co-speech gestures, we introduce various metrics:
\begin{itemize}[leftmargin=*]
    \vspace{-0.6em}
    \item \textbf{FGD}: Fréchet Gesture Distance (FGD) \cite{yoon2020speech} is calculated as the distribution distance between the body movements of synthesized results and real ones via a pre-trained autoencoder.
    \item \textbf{BC}: Beat Consistent Score (BC) / Beat Alignment Score (BA) \cite{liu2022beat, Liu_2024_CVPR} measures whether the generated motion dynamics are rhythmic consistent with the input speech audios. We report the average score of two speakers in our experiments.
    \item \textbf{Diversity}: Similar to \citep{liu2022learning, zhu2023taming, Liu_2024_CVPR}, the autoencoder of FGD is exploited to acquire feature embeddings of the synthesized gestures. Here, the diversity score means the average distance of 500 randomly assembed pairs.
\end{itemize}

\subsection{Quantitative Results}

\begin{table}[t]
\centering
\caption{Comparison with the state-of-the-art counterparts on our newly collected GES-Inter dataset. $\uparrow$ means the higher the better,  and $\downarrow$ indicates the lower the better. $\pm$ means 95\% confidence interval. The dotted line separates whether the methods are adopted from single-person co-speech generation or text2motion counterparts.}
\label{tab:comparison}
\footnotesize
\setlength{\tabcolsep}{6 mm}
\begin{tabular}{lccc}
\toprule
 & \multicolumn{3}{c}{GES-Inter Dataset} \\ \cmidrule(r){2-4} 
\multirow{-2}{*}{Methods} & FGD $\downarrow$ & BC $\uparrow$ & Diversity $\uparrow$ \\ \midrule \midrule
TalkSHOW~\citep{yi2023generating}\textcolor[HTML]{C0C0C0}{$_{CVPR}$} & 2.256 & 0.613 & 53.037$^{\pm1.021}$ \\
ProbTalk~\citep{liu2024towards}\textcolor[HTML]{C0C0C0}{$_{CVPR}$} & 1.238 & 0.645 & 46.981$^{\pm2.173}$ \\
DiffSHEG~\citep{chen2024diffsheg}\textcolor[HTML]{C0C0C0}{$_{CVPR}$} & 1.209 & 0.638 & 56.781$^{\pm1.905}$ \\
EMAGE~\citep{Liu_2024_CVPR}\textcolor[HTML]{C0C0C0}{$_{CVPR}$} & 1.884 & 0.637 & 60.917$^{\pm1.179}$ \\ \hline
MDM~\citep{tevet2023human}\textcolor[HTML]{C0C0C0}{$_{ICLR}$} & 1.696 & 0.654 & 65.529$^{\pm2.218}$ \\
InterX~\citep{xu2024inter}\textcolor[HTML]{C0C0C0}{$_{CVPR}$} & 1.178 & 0.661 & 65.161$^{\pm1.010}$ \\
InterGen~\citep{liang2024intergen}\textcolor[HTML]{C0C0C0}{$_{IJCV}$} & 1.012 & 0.670 & 69.455$^{\pm1.590}$ \\ \midrule
\rowcolor[HTML]{ECF4FF} 
\textbf{\modelname (ours)} & \textbf{0.769} & \textbf{0.692} & \textbf{72.824$^{\pm2.026}$} \\ \bottomrule
\end{tabular}%
\vspace{-1em}
\end{table}

\noindent{\textbf{Comparisons with SOTA Methods.}}
To the best of our knowledge, we are the first to explore the coherent concurrent co-speech gesture generation with conversational human audio. To fully verify the superiority of our method, we implement various state-of-the-art (SOTA) counterparts from the perspective of single-person-based co-speech gesture generation (\ie, TalkSHOW~\citep{yi2023generating}, ProbTalk~\citep{liu2024towards}, DiffSHEG~\citep{chen2024diffsheg}, EMAGE~\citep{Liu_2024_CVPR}) and text2motion generation (\ie, MDM~\citep{tevet2023human}, InterX~\citep{xu2024inter} InterGen~\citep{liang2024intergen}). For fair comparisons, all the competitors are implemented by official source codes or pre-trained models released by authors. Specifically, in DiffSHEG, we follow the convention of the original work to utilize the pre-trained HuBERT~\citep{hsu2021hubert} for audio feature extraction. In TalkSHOW, we exploit the pre-trained Wav2vec~\citep{baevski2020wav2vec} to encode the audio signals following the original setting. Apart from this, the remaining components for gesture generation in DiffSHEG and TalkSHOW are all trained from scratch on the newly collected GES-Inter dataset. For other methods, we modify their final output layer to match the dimensions of our experimental settings.
Since the above text2motion counterparts are designed without the audio incorporation setting, we adopt the same audio encoder as ours in the models.

As reported in Table~\ref{tab:comparison}, we adopt the FGD, BC, and diversity for a well-rounded view of comparison. Our \modelname outperforms all the competitors by a large margin on the GES-Inter dataset. Remarkably, our method even achieves more than 24\% (\ie, $(1.102-0.769)/1.012 \approx 24\%$) improvement over the sub-optimal counterpart in FGD. We observe that both InterGen~\citep{liang2024intergen} and ours synthesize the authority gestures with much higher diversity than others. This is caused by both of them employing the bilateral branches to generate concurrent gestures of two speakers. However, InterGen shows lower performance on FGD due to the lack of effective temporal interaction modeling. In terms of BC, our method attains much better results than other counterparts. This aligns highly with our insight on the audio separation-based bilateral diffusion backbone that encourages each branch to synthesize speech coherent gestures while preserving the vivid interaction of the results. Compared with single-person co-speech gesture based ones, our model still achieves the best performance. This can be attributed to our well-designed temporal interaction module. 

\paragraph{Ablation Study.}
To further evaluate the effectiveness of our \modelname, we conduct extensive ablation studies of different components and experiment settings as variations.

\noindent{\textbf{Effects of the TIM and Mutual Attention:}} To verify the effectiveness of TIM and mutual attention mechanism, we conduct detailed experiments as reported in Table~\ref{tab:ablation1}. The exclusion of the temporal interaction module (TIM) and the mutual attention mechanism lead to performance degradation in our full-version framework, respectively. Moreover, we conduct ablation by simply replacing the TIM with an MLP layer for feature fusion, the FGD and BC display the obvious worse impact as shown in the Table. The results verify that our TIM effectively enhances interactive coherency between two speakers. In particular, our temporal interaction module effectively models the temporal dependency between the gesture motions of the current speaker and the conversation partner. 
\begin{wraptable}{r}{0.6\textwidth}
\centering
\caption{Ablation study of TIM and mutual attention mechanism on our GES-Inter dataset.}
\label{tab:ablation1}
\setlength{\tabcolsep}{2 mm}{%
\footnotesize
\begin{tabular}{lccc}
\toprule
 & \multicolumn{3}{c}{GES-Inter Dataset} \\ \cmidrule{2-4} 
\multirow{-2}{*}{Methods} & FGD $\downarrow$ & BC $\uparrow$ & Diversity $\uparrow$ \\ \midrule \midrule
w/ MLP & 1.202 & 0.663 & 64.690$^{\pm1.137}$ \\
w/o TIM & 1.297 & 0.676 & 67.953$^{\pm1.203}$ \\
w/o Mutual Attention & 0.924 & 0.681 & 69.084$^{\pm1.412}$ \\ \midrule
\rowcolor[HTML]{ECF4FF} 
\textbf{\modelname (full version)} & \textbf{0.769} & \textbf{0.692} & \textbf{72.824$^{\pm2.026}$} \\ \bottomrule
\end{tabular}%
}
\end{wraptable}
Therefore, implementation without it leads our framework to fail in producing cooperative motions, thus significantly reducing the performance in all the metrics.
Moreover, the exclusion of the mutual attention mechanism results in FGD is obviously worse than the full version framework. This indicates that our mutual attention can effectively handle complex interactions from the perspective of holistic fashion while balancing the specific movements of two speakers.

\noindent{\textbf{Effects of the Bilateral Branches and Audio Mixed/ Separation:}} 
We also conduct the ablation study in different experiment settings. To demonstrate the effectiveness of our bilateral diffusion branches, we construct the single branch-based diffusion pipeline that generates the gestures of two speakers in a holistic manner. As shown in Table~\ref{tab:ablation2}, by subtracting the bilateral branches from the full version pipeline, the indicator FGD displays much worse results (\eg, $0.769 \to 1.669$).
This outcome verifies that our cooperative bilateral diffusion branches effectively handle the asymmetric motion of concurrent gestures of two speakers. This supports our key technical insight on framework construction. Then, by subtracting the original mixed audio, the indicators FGD and BC present much worse performance. These results verify the mixed audio signal displays effectively enhance the interaction between two speakers. 
\begin{wraptable}{r}{0.6\textwidth}
\centering
\vspace{-1em}
\caption{Ablation study of bilateral branches and audio mixed/ separation on our GES-Inter dataset.}
\label{tab:ablation2}
\setlength{\tabcolsep}{2 mm}{%
\footnotesize
\begin{tabular}{lccc}
\toprule
 & \multicolumn{3}{c}{GES-Inter Dataset} \\ \cmidrule{2-4} 
\multirow{-2}{*}{Methods} & FGD $\downarrow$ & BC $\uparrow$ & Diversity $\uparrow$ \\ \midrule \midrule
w/o Bilateral Branches & 1.669 & 0.640 & 64.542$^{\pm1.252}$ \\
w/o Mixed Audio & 1.227 & 0.656 & 64.899$^{\pm1.004}$ \\
w/o Audio Separation & 1.180 & 0.633 & 66.159$^{\pm1.501}$ \\ \midrule
\rowcolor[HTML]{ECF4FF} 
\textbf{\modelname (full version)} & \textbf{0.769} & \textbf{0.692} & \textbf{72.824$^{\pm2.026}$} \\ \bottomrule
\end{tabular}%
}
\end{wraptable}
Meanwhile, we further verify the capability of our audio separation design where the model only takes the mixed conversational speech signal as input. Based on the above-mentioned fashion, the BC metric clearly attains a much worse result than the separated one.
Directly modeling mixed conversational speech to produce concurrent gestures impacts the interactive correlation between two speakers. This seriously affects the harmony of synthesized concurrent gestures and corresponding speech rhythm.

\noindent{\textbf{Effects of the Foot Contact Loss:}} Inspired by \citep{tevet2023human, liang2024intergen}, we introduce foot contact loss to ensure the physical reasonableness of the generated gestures.
\begin{wraptable}{r}{0.6\textwidth}
\centering
\vspace{-1em}
\caption{Ablation study of foot contact loss on our GES-Inter dataset.}
\label{tab:ablation3}
\setlength{\tabcolsep}{2 mm}{%
\footnotesize
\begin{tabular}{lccc}
\toprule
 & \multicolumn{3}{c}{GES-Inter Dataset} \\ \cmidrule{2-4} 
\multirow{-2}{*}{Methods} & FGD $\downarrow$ & BC $\uparrow$ & Diversity $\uparrow$ \\ \midrule \midrule
w/o Foot Contact $\mathcal{L}_{foot}$ & 1.082 & 0.675 & 68.448$^{\pm1.082}$ \\ \midrule
\rowcolor[HTML]{ECF4FF} 
\textbf{\modelname (full version)} & \textbf{0.769} & \textbf{0.692} & \textbf{72.824$^{\pm2.026}$} \\ \bottomrule
\end{tabular}%
}
\vspace{-0.5em}
\end{wraptable}
Since we only model the upper body joints in experiments, we complete the lower body joints as $T$ pose in forward kinematic function during calculate loss. We conduct the ablation study to verify the effectiveness of $\mathcal{L}_{foot}$. 
As illustrated in Table~\ref{tab:ablation3}, the exclusion of the foot contact loss results in FGD and BC are obviously worse than the full version framework. This indicates that our foot contact loss displays a positive impact on the generated postures.

\vspace{-1em}
\subsection{Qualitative Evaluation}
\begin{figure}[t]
\begin{center}
\includegraphics[width=1\linewidth]{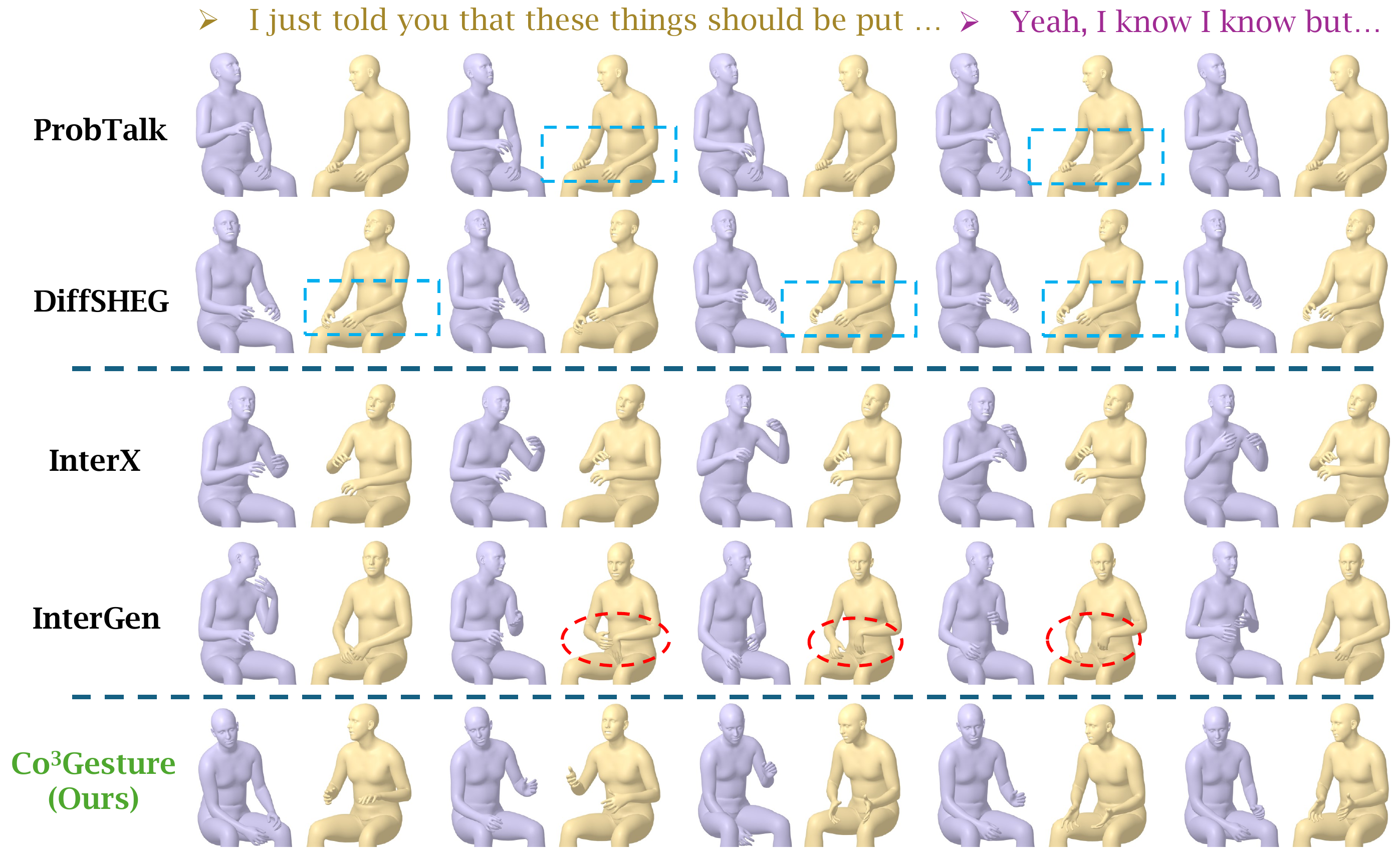}
\end{center}
\caption{ Visualization of our generated concurrent 3D co-speech gestures against various state-of-the-art methods. The samples are from our newly collected GES-Inter dataset. 
}
\label{fig:comparison}
\vspace{-1em}
\end{figure}
\vspace{-0.5em}

\paragraph{Visualization.}
To fully demonstrate the superior performance of our method, we display the visualized keyframes generated by our \modelname framework with other counterparts, as illustrated in Figure~\ref{fig:comparison}.
For better demonstration, the relative position coordinates of the two speakers are fixed in visualization. The lower body including the legs is fixed (\eg, seated) while visualizing due to the weak correlation with human speech. For example, it is quite challenging to model whether the two speakers are sitting or standing from only audio inputs.
We showcase the two optimal methods from single-person gesture generation and text2motion, respectively. Our method shows coherent and interactive body movements against other ones. To be specific, we observe that ProbTalk and DiffSHEG would synthesize the stiff results (\eg, blue rectangles of right speakers). Although the Inter-X generates the natural movements of the left speaker, it displays less interactive dynamics of the right speaker. In addition, the results synthesized by InterGen show reasonable interaction between two speakers. However, it may produce unnatural postures sometimes (as depicted in red circles). In contrast, our \modelname can generate interaction coherent concurrent co-speech gestures. This highly aligns with our insight into the bilateral cooperative diffusion pipeline. For more visualization demo videos please refer to our anonymous webpage: \href{https://mattie-e.github.io/Co3/}{\textit{https://mattie-e.github.io/Co3/}}. In our experiments, the length of the generated gesture sequence is 90 frames with 15 FPS. Thus, all the demo videos in the user study have the same length of 6 seconds.

\begin{wrapfigure}{r}{0.5\textwidth}
    \centering
    \vspace{-0.7em}
\includegraphics[width=0.5\textwidth]{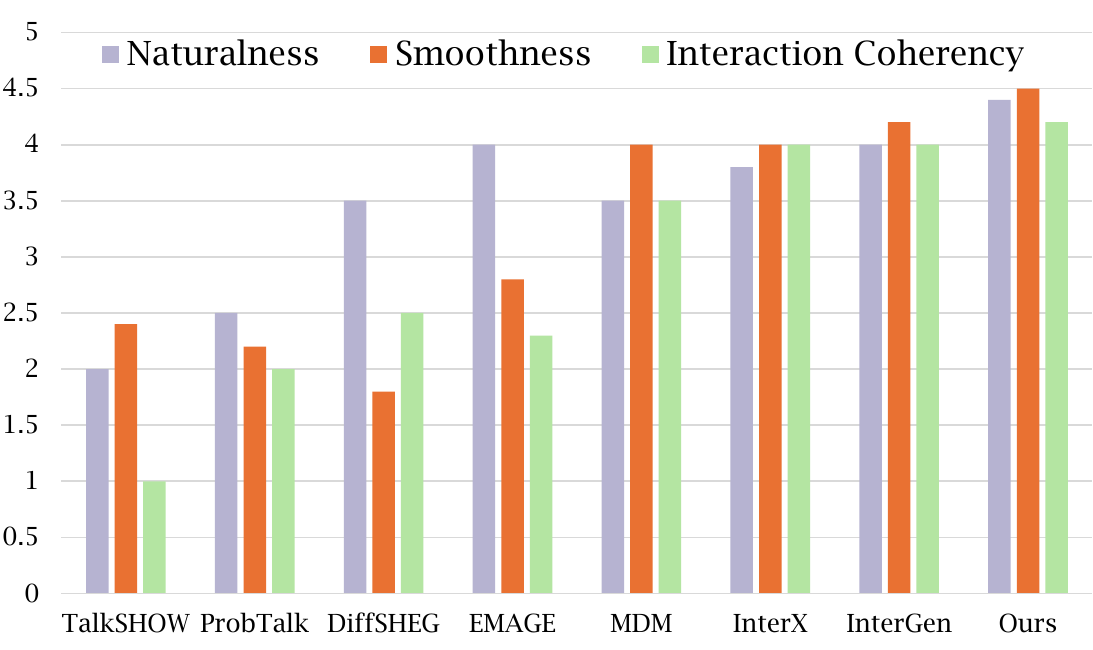}
\caption{User study on gesture naturalness, motion smoothness, and interaction coherency.}
\vspace{-2em}
    \label{fig:user_study}
\end{wrapfigure}

\paragraph{User Study.}
To further analyze the quality of concurrent gestures synthesized by ours against various competitors, we conduct a user study by recruiting 15 volunteers. 
All the volunteers are anonymously selected from various majors in school. For each method, we randomly select two generated videos in the user study. Hence, each participant needs to watch 16 videos for 6 seconds of each.
The subjects are required to evaluate the generated results by all the counterparts in terms of naturalness, smoothness, and interaction coherency. The visualized videos are randomly selected and ensure that each method has at least two samples. The statistical results are shown in Figure~\ref{fig:user_study} with the rating scale from 0-5 (the higher, the better). 
Our framework demonstrates the best performance compared with all the competitors. To be specific, our method achieves noticeable advantages from the perspective of smoothness and interaction coherency. 
This indicates the effectiveness of our proposed bilateral denoising and temporal interaction module.

%% file: section/5_conclusion.tex
\section{Conclusion}
\vspace{-0.5em}
In this paper, we introduce a new task of coherent concurrent co-speech gesture generation given conversational human speech. We first newly collected a large-scale dataset containing more than 7M concurrent co-speech gesture instances of two speakers, dubbed GES-Inter. This high-quality dataset supports our task while significantly facilitating the research on 3D human motion modeling. Moreover, we propose a novel framework named \modelname that includes a temporal-interaction module to ensure the generated gestures preserve interactive coherence. Extensive experiments conducted on our GES-Inter dataset show the superiority of our framework.
\paragraph{Limitation.}
Despite the huge effort we put into data preprocessing, the automatic pose extraction stream may influence our dataset with some bad instances. Meanwhile, our framework only generates the upper body movements without expressive facial components.  In the future, we will incorporate our framework with tailor-designed facial expression modeling and investigate more stable data collection techniques to further improve the quality of our dataset. Besides, we will put more effort into designing specific interaction metrics for better concurrent gesture evaluation.

\section*{Acknowledgment}
The research was supported by Early Career Scheme (ECS-HKUST22201322), Theme-based Research Scheme (T45-205/21-N) from Hong Kong RGC, NSFC (No. 62206234), and Generative AI Research and Development Centre from InnoHK.

%% file: section/6_appendix.tex
\appendix
\section{Appendix}
To showcase the superior quality of our GES-Inter dataset and the effectiveness of our proposed \modelname, we provide additional details on data collection and further visualization results below.

\begin{figure}
\begin{center}
\includegraphics[width=1\linewidth]{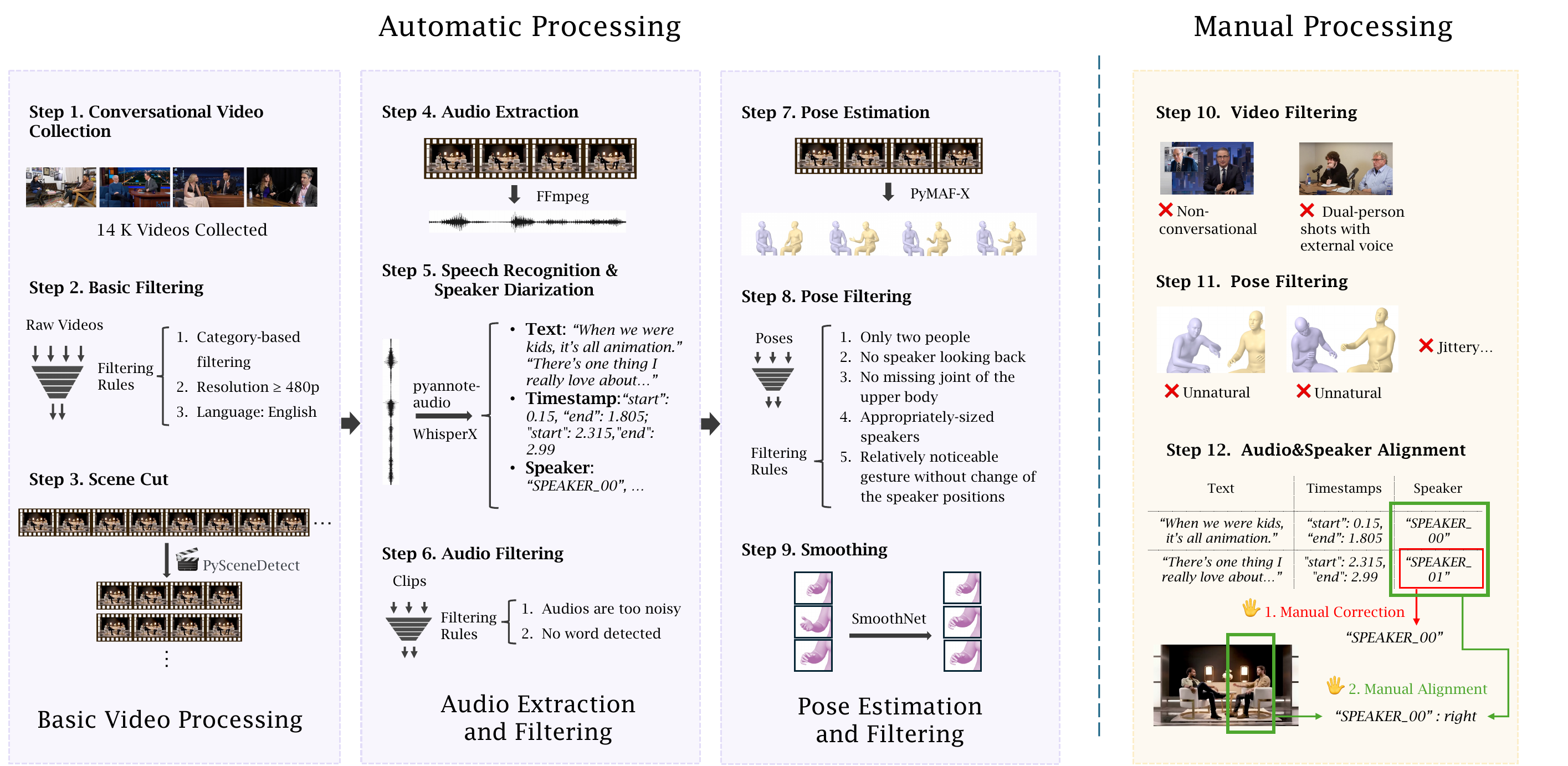}
\end{center}
\caption{The overall workflow of our dataset construction. The videos are processed to obtain high-quality postures through advanced automatic technologies and professional expert proofreading.
}
\label{fig:data_pipeline}
\end{figure}

\subsection{Dataset Construction}
In this section, we give a detailed explanation of the data processing pipeline of our GES-Inter dataset. We summarize the acquisition, processing, and filtering of our GES-Inter dataset into two main procedures: automatic and manual processing steps, as illustrated in Figure~\ref{fig:data_pipeline}.

\subsubsection{Automatic Processing Steps}
To build a high-quality 3D co-speech gesture dataset with concurrent and interactive body dynamics, we collect a considerable number of videos. They are then processed using automated methods to extract both audio and motion information.

\noindent{\textbf{Basic Video Processing (Step 1, 2, 3):}} 
First, with related searching keywords, we collect more than 14K conversational videos along with their metadata (\eg, video length, frame resolution, audio sampling rate, \etc). Those keywords include talk show, conversation, interview, \etc
In this step, we acquire raw video data totaling up to 1,095 hours. However, many of these videos do not meet our requirements regarding category, quality, language, and other factors.
Therefore, we filter them in step 2 to make sure: i) the speakers in the videos are real people rather than cartoon characters; ii) the videos meet an acceptable quality standard, featuring a resolution of at least 480p for clear visuals; and iii) only English conversations are included. In these preprocessing phases, due to the large amount of raw video collected and labor consumption, we conduct initial filtering using automatic techniques without manually checking each video. Specifically, we leverage YOLOv8 for human detection, discarding clips that do not show realistic people (eg, cartoon characters). Meta information provided by downloaded videos directly filters the English conversational corpus. Following the initial filtering, we proceed to step 3, where we use PySceneDetect to cut the videos into short clips. 

\noindent{\textbf{Audio Extraction and Filtering (Step 4, 5, 6):}}
Audios and poses are two necessary attributes for our GES-Inter dataset. In Step 4, we extract audio from the video clips using FFmpeg. 
In step 5, we initially employed the pyannote-audio technique for speaker diarization, configuring it for two speakers to accommodate two-person dialogues. The pyannote-audio tool assigns each speech segment to the appropriate speaker. Next, we utilize WhisperX~\cite{bain2022whisperx} for speech-to-text transcription. After transcription, we cluster the speakers based on the generated timestamps to better organize the dialogue.
With the extracted audio and speaker diarization, we filter out clips in Step 6 that either have relatively low audio quality or no word detected. This filtering process improves the efficiency of the subsequent pose estimation.

\noindent{\textbf{Pose Estimation and Filtering (Step 7, 8, 9):}} 
As an acknowledged 3D human representation standard, SMPL-X ~\cite{pavlakos2019expressive} is used to represent whole-body poses in various related tasks ~\cite{jiang2023motiongpt, zhang2022motiondiffuse, liu2024towards, Liu_2024_CVPR, chen2024diffsheg, liang2024intergen}. Accordingly, we employ the cutting-edge pose estimator PyMAF-X ~\cite{zhang2023pymaf} to extract high-quality 3D postures including body poses, subtle fingers, shapes, and expressions of the speakers. We then apply five criteria to filter the clips based on the pose annotation: \textit{containing only two people, no speaker looking back, no missing joint of the upper body, appropriately-sized speakers, and relatively noticeable gesture without change of the speaker positions}. However, upon examining the visualized motions, we still observe that some temporal jittering within the movements is inevitable. To this end, we exploit SmoothNet~\cite{zeng2022smoothnet} for temporal smoothing and jittery motion refinement in Step 9. The jittery effects are mostly caused by the blurring of speakers moving quickly in consecutive video frames. Due to the strict keyword selection in raw video crawling, our dataset rarely contains two speakers standing or walking around. If there are several clips including the aforementioned postures, we will filter them out to ensure our dataset maintains unified posture representation.

In particular, our manual review indicates that SmoothNet effectively generates cleaner and more reliable motion sequences while maintaining a diverse range of postures. However, due to the frequent extreme variations in camera angles, speaker poses, and lighting in talk show videos, some inaccuracies in pose estimations from PyMAF-X are unavoidable. Thus, inspired by \cite{pavlakos2019expressive}, we translate the joint of arm postures as Euler angles with $x$, $y$, and $z$ order. Then, if the wrist poses exceed 150 degrees on any axis, or if the pose changes by more than 25 degrees between adjacent frames (at 15 fps), we discard these abnormal postures over a span of 90 frames.

\subsubsection{Manual Processing Steps}
Building on the initial postures and audio obtained automatically, we introduce manual processing in this section to further refine the annotation.

\noindent{\textbf{Basic Video Filtering (Step 10):}} 
We observe that several undesired clips have passed through the initial filtering, including non-conversational scenarios and dual-person shots with external voices. To ensure that all videos meet our standards, we recruit two groups of inspectors to meticulously review and eliminate any that do not comply with the specified criteria. The results of each group are sampled and inspected to guarantee authority.

\noindent{\textbf{Pose Filtering (Step 11):}} 
We conduct a manual review of the processed clips at a consistent ratio of 5:1, selecting one clip from each group of five while adhering to the order of scenecut. This approach is valid, as there may exist overlap among adjacent clips. 
All clips are organized into five groups, and each group is assigned to an inspector for thorough evaluation. The inspectors assess the visualizations using SMPL-X parameters to determine whether the motion appears smooth, jittery, or unnatural. If any sequences are identified as jittery or unnatural, we discard the entire group of five clips from which the sample was taken. 

Additionally, we eliminate instances where the speakers experience significant occlusion of their bodies during the interaction. This meticulous evaluation process greatly enhances the quality of our GES-Inter dataset.

\noindent{\textbf{Audio\&Speaker Alignment (Step 12):}}
We obtained speech separation results with an accuracy of 95\%. Here we defined correct instances as those audio clips with accurate speech segmentation, correct text recognition, and accurate alignment. During our audio preprocessing, the audio is initially segmented by pyannote-audio technique to achieve 92\% accuracy. Then, the accuracy of text recognized by WhisperX is 96\%.

Once we obtain the separated audios, to ensure the identity consistency between the separated audio and body dynamics, we conduct audio-speaker alignment in this step. To be specific, professional human inspectors are recruited to manually execute this operation.
Inspectors first check every video clip with its diarization to ensure the sentence-level speaker identities are correct and consistent within the clip. Then, inspectors assign the specific spatial position, \ie, left or right, to the speaker identities in the diarization. Using the alignment and the revised diarization with timestamps, inspectors separate each extracted audio into two distinct files and name them according to the corresponding speakers. To ensure the high fidelity of the alignment, the initially aligned audio-speaker pairs are double-checked by another group of inspectors. 
Meanwhile, the human inspectors would further check the rationality of segmentation and text recognition results from the perspective of human perception. In this step, we set two groups of inspectors for cross-validation to ensure the final alignment rate is 98\%.
In this manner, our GES-Inter dataset contains high-quality human postures with corresponding separated authority human speeches and multi-modality annotations. We provide examples of audio separation for better demonstration (refer to our webpage: \href{https://mattie-e.github.io/Co3/}{\textit{https://mattie-e.github.io/Co3/}}).

\subsection{More Details about Experimental Setting}
Due to the complex and variable positions of the two speakers of in-the-wild videos, we set the relative positions of the two speakers to fixed values. In the experiments, we only model the upper body dynamics of the two speakers. In particular, the joint order follows the convention of SMPL-X. During experiments, we follow the convention of \citep{liu2022beat,liu2022learning, Liu_2024_CVPR} to resample FPS as 15. In our dataset, we retain all the metadata (\eg, video frames, poses, facial expressions) within the original FPS (\ie, 30) of talk show videos. We will release our full version data and pre-processing code, thereby researchers can obtain various FPS data according to their tasks.

\subsection{More Details about User Study}
In the user study, all participating students are asked to evaluate each video without any indication of which model generated it. For fair comparison in user study, the demo videos are randomly selected. We count the motion fractions length of two speakers upon all the 16 demo videos. We adopt elbow joints as indicators to determine whether the motion occurs. Empirically, when the pose changes by more than 5 degrees between adjacent frames, we nominate the speakers who are moving now. Therefore, among 16 demo videos with 6 seconds, the average motion fraction lengths of the two speakers are 4.3 and 3.1 seconds.

A higher score reflects better quality, with 5 signifying that the video fully meets the audience's expectations, while 0 indicates that the video is completely unacceptable. To ensure fairness, each video is presented on a PowerPoint slide with a neutral background. Before participants see the generated results, we show several pseudo-annotated demos in our dataset as reference. All participants are required to watch the video at least once before they rate it. We invite participants in batches at different time periods within a week. 
Once all students have submitted their ratings anonymously, we collect them to calculate an average score. After completing the statistics, we randomly selected 60\% of them to rate again two weeks later, and the results show that there is no obvious deviation. 

We report the detailed mean and standard deviation for each method as shown in Table~\ref{tab:data_user_study}. Our method even achieves a 10\% ((4.4-4.0)/4=10\%) large marginal improvement over suboptimal InterGen in Naturalness. Meanwhile, our method displays a much lower standard deviation than InterX and InterGen. This indicates the much more stable performance of our method against competitors.

\begin{table}[]
\centering
\caption{Statistical results in User Study. $\pm$ denotes standard deviation.}
\label{tab:data_user_study}
\footnotesize
\begin{tabular}{cccc}
\toprule
Comparison Methods & Naturalness & Smoothness & Interaction Coherency \\ \midrule \midrule
TalkSHOW & 2$^{\pm 0.1}$ & 2.4$^{\pm 0.6}$ & 1$^{\pm 0.1}$ \\
ProbTalk & 2.5$^{\pm 0.3}$ & 2.2$^{\pm 0.3}$ & 2$^{\pm 0.2}$ \\
DiffSHEG & 3.5$^{\pm 0.5}$ & 1.8$^{\pm 0.3}$ & 2.5$^{\pm 0.6}$ \\
EMAGE & 4$^{\pm 0.4}$ & 2.8$^{\pm 0.4}$ & 2.3$^{\pm 0.5}$ \\
MDM & 3.5$^{\pm 0.6}$ & 4$^{\pm 0.3}$ & 3.5$^{\pm 0.1}$ \\
InterX & 3.8$^{\pm 0.4}$ & 4$^{\pm 0.5}$ & 4$^{\pm 0.3}$ \\
InterGen & 4$^{\pm 0.5}$ & 4.2$^{\pm 0.2}$ & 4$^{\pm 0.2}$ \\ \midrule
\rowcolor[HTML]{ECF4FF} 
Ours & 4.4$^{\pm 0.2}$ & 4.5$^{\pm 0.1}$ & 4.2$^{\pm 0.1}$ \\ \bottomrule
\end{tabular}
\end{table}

To further verify the effectiveness of our user study, we conduct a significant analysis of the user study using t-test, focusing on three key aspects: Naturalness, Smoothness, and Interaction Coherency. The results verify our method surpasses all the counterparts with significant improvements, including sub-optimal InterGen. In particular, for all the comparisons between our model and the other models, we formulate our null hypotheses (H0) as "our model does not outperform another method". In contrast, the alternative hypothesis (H1) posits that "our model significantly outperforms another method," with a significance level ($\alpha$) set as 0.05. Here, we perform a series of t-tests to compare the rating scores of our model against each of the other competitors individually and calculate all the t-statistics shown in Table~\ref{tab:user_study}. Since we recruit 15 volunteers, our degree of freedom(df) for every analysis is 14. Then, we look up the t-table with two tails and find out all the p-values are less than 0.05 ($\alpha$). Therefore, we reject the null hypotheses, indicating that our model significantly outperforms all the other methods in every aspect.

\begin{table}[t]
\centering
\caption{Significance Analysis of User Study}
\label{tab:user_study}
\footnotesize
\begin{tabular}{cccc}
\toprule
Comparison Methods & Naturalness & Smoothness & Interaction Coherency \\ \midrule \midrule
TalkSHOW           & 5.345       & 3.567      & 4.123                 \\
ProbTalk           & 3.789       & 5.001      & 3.456                 \\
DiffSHEG           & 4.789       & 2.654      & 5.299                 \\
EMAGE              & 3.214       & 5.120      & 4.789                 \\
MDM                & 3.789       & 4.567      & 2.987                 \\
InterX             & 3.001       & 3.456      & 2.148                 \\
InterGen           & 2.654       & 3.299      & 2.234                 \\ \bottomrule
\end{tabular}
\end{table}

\subsection{Additional Visualization Results}
\begin{figure}[t]
\begin{center}
\includegraphics[width=1\linewidth]{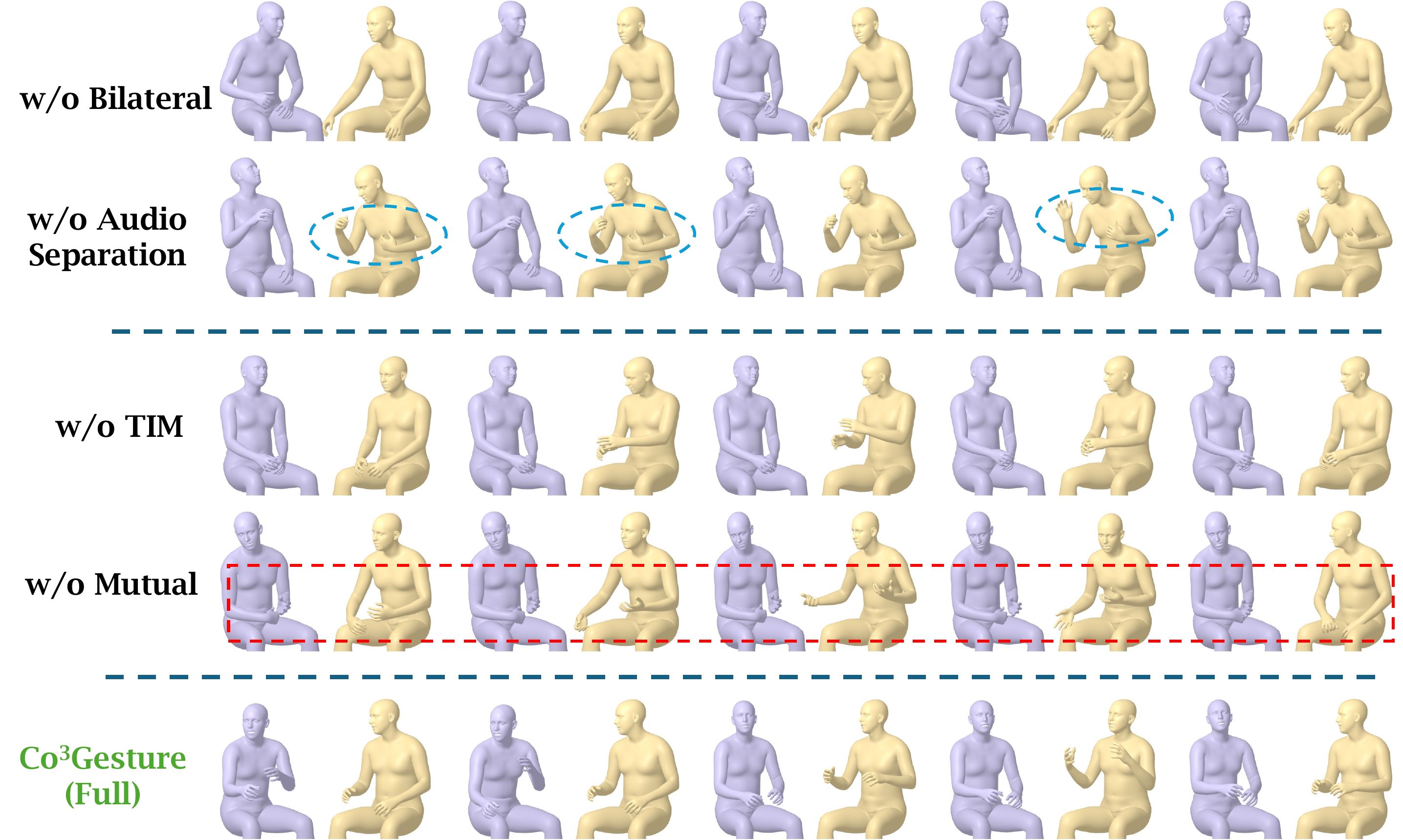}
\end{center}
\vspace{-0.5em}
\caption{ Visualization of our generated concurrent 3D co-speech gestures in the ablation study. Best view on screen.
}
\label{fig:ablation}
\end{figure}

Here, we provide more visualization results of the ablation study in our experiments. As shown in Figure~\ref{fig:ablation}, the full version of our framework demonstrates the vivid and coherent interaction of body dynamics against other versions. We also display more visualized demo videos on our anonymous website: \href{https://mattie-e.github.io/Co3/}{\textit{https://mattie-e.github.io/Co3/}}.

